\documentclass[letterpaper]{article} 
\usepackage[submission]{aaai23}  
\usepackage{times}  
\usepackage{helvet}  
\usepackage{courier}  
\usepackage[hyphens]{url}  
\usepackage{graphicx} 
\usepackage{mathrsfs}
\usepackage{amsmath}
\usepackage{algorithm}
\usepackage{algorithmic}
\usepackage{multicol}
\usepackage{multirow}
\usepackage{amsfonts}
\usepackage{makecell}
\usepackage{natbib}  
\usepackage{caption} 
\usepackage{algorithm}
\usepackage{algorithmic}

\usepackage[toc,page,header]{appendix}

\usepackage{newfloat}
\usepackage{listings}
\DeclareCaptionStyle{ruled}{labelfont=normalfont,labelsep=colon,strut=off} 
\floatstyle{ruled}
\newfloat{listing}{tb}{lst}{}
\floatname{listing}{Listing}
\title{Layout-aware Dreamer for Embodied Referring Expression Grounding}
\author {
    Mingxiao Li\equalcontrib,\textsuperscript{\rm 1}
    Zehao Wang\equalcontrib,\textsuperscript{\rm 2}
    Tinne Tuytelaars, \textsuperscript{\rm 2}
    Marie-Francine Moens, \textsuperscript{\rm 1}
}
\affiliations {
    \textsuperscript{\rm 1} Computer Science Department of KU Leuven \\
    \textsuperscript{\rm 2} Electrical Engineering Department (ESAT-PSI) of
        KU Leuven \\ 
    mingxiao.li@cs.kuleuven.be \\ 
    zehao.wang@esat.kuleuven.be
}

\usepackage{bibentry}

\begin{document}

\maketitle

\begin{abstract}
In this work, we study the problem of Embodied Referring Expression Grounding, where an agent needs to navigate in a previously unseen environment and to localize a remote object described by a concise high-level natural language instruction. When facing such a situation, a human tends to imagine what the destination may look like and to explore the environment based on prior knowledge of the environmental layout, such as the fact that a bathroom is more likely to be found near a bedroom than a kitchen. We have designed an autonomous agent called Layout-aware Dreamer (LAD), including two novel modules, that is, the \textit{Layout Learner} and the \textit{Goal Dreamer} to mimic this cognitive decision process. The \textit{Layout Learner} learns to infer the room category distribution of neighboring unexplored areas along the path for coarse layout estimation, which effectively introduces layout common sense of room-to-room transitions to our agent. To learn an effective exploration of the environment, the \textit{Goal Dreamer} imagines the destination beforehand. Our agent achieves new state-of-the-art performance on the public leaderboard of the REVERIE dataset in challenging unseen test environments with improvement in navigation success (SR) by $4.02\%$ and remote grounding success (RGS) by $3.43\%$ 
compared to the previous state-of-the-art. The code is released at \url{https://github.com/zehao-wang/LAD}
\end{abstract}

\section{Introduction}

In recent years, embodied AI has matured. In particular, a lot of works~\cite{vlnduet,hao2020towards,majumdar2020improving,reinforced,song2022one,vlnce_topo,georgakis2022cm2} have shown promising results in Vision-and-Language Navigation (VLN)~\cite{room2room,vlnce}. In VLN, an agent is required to reach a destination following a fine-grained natural language instruction that provides detailed step-by-step information along the path, for example ``Walk forward and take a right turn. Enter the bedroom and stop at the bedside table". However, in real-world applications and human-machine interactions, it is tedious for people to give such detailed step-by-step instructions. Instead, a high-level instruction only describing the destination, such as ``Go to the bedroom and clean the picture on the wall.", is more usual. 
%

In this paper, we target 
such high-level instruction-guided tasks. Specifically, we focus on the Embodied Referring Expression Grounding task~\cite{reverie,zhu2021soon}. In this task, an agent receives a high-level instruction referring to a remote object, and it needs to explore the environment and localize the target object. When given a high-level instruction, 
we humans tend to imagine what the scene of the destination looks like. Moreover,  we can  efficiently navigate to the target room, even in previously unseen environments, by exploiting commonsense knowledge about the layout of the environment.
However, for an autonomous agent, generalization to unseen environments still remains challenging. 

\begin{figure}[t]
    \centering
    \includegraphics[width=\linewidth]{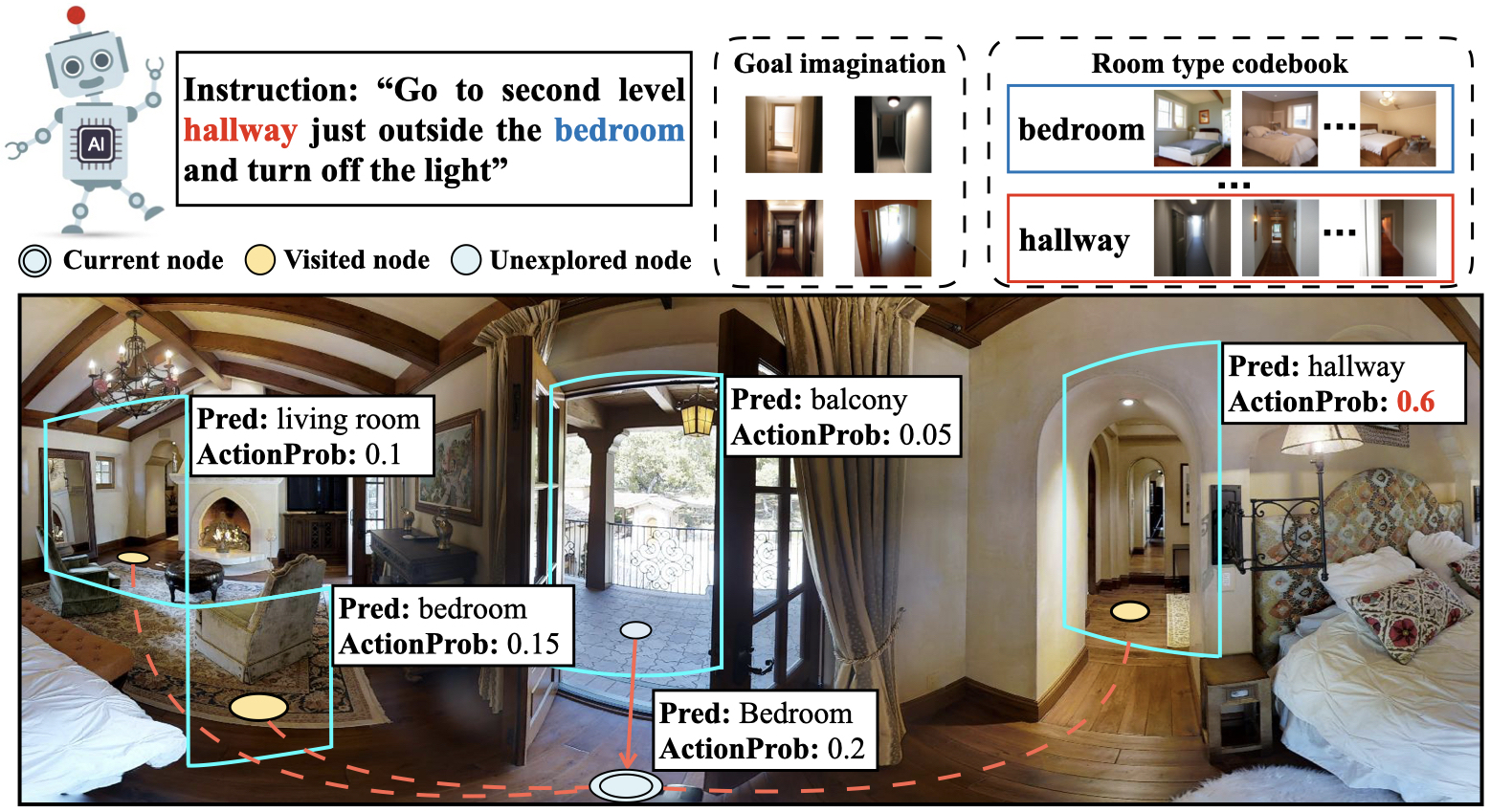}
    \caption{The agent is required to navigate and find the mentioned object in the environment. Based on acquired commonsense knowledge, the agent correctly identifies the current and surrounding room types. Based on the imagination of the destination, it correctly chooses the unexplored right yellow dot as the next step to go.}
    \label{fig: tessar}
\end{figure}

Inspired by how humans make decisions when receiving high-level instructions in unseen environments, 
as shown in Fig.~\ref{fig: tessar}, we design an agent that can identify the room type of current and neighboring navigable areas based on a room type codebook and previous states. On top of that, it learns to combine this information with goal imagination to jointly infer the most probable moving direction.
Thus, we propose two modules named \textbf{Layout Learner} and \textbf{Goal Dreamer} to achieve this goal.
%
In our model, the agent stores trajectory information by building a real-time topological map, where nodes represent either visited or unexplored but partially visible areas. The constructed topological map can be seen as a long-term scene memory. At each time step, the node representation is updated by moving the agent to the current node and receiving new observations. 
%
The \textbf{Layout Learner} module learns to infer the layout distribution of the environment with a room-type codebook constructed using a large-scale pre-trained text-to-image model, GLIDE~\cite{glide} in our case. The codebook design helps the model to leverage high-level visual commonsense knowledge of room types and boosts performance in layout learning. This prediction is updated at each time step allowing the agent to correct its prediction when more areas are explored. In the \textbf{Goal Dreamer} module, we encourage the agent to imagine the destination beforehand by generating a set of images with the text-to-image model GLIDE. The use of this imagination prior helps accurate action prediction. The cross-attention between topological node representations and imagination features is conducted, and its output is used to help the agent make a decision at each time step. 
In summary, the contributions of our paper are threefold:
\begin{itemize}
    \item {We propose a \textbf{Layout Learner} which leverages the visual commonsense knowledge from a room-type codebook generated using the GLIDE model. It not only helps the agent to implicitly learn the environment layout distribution but also to better generalize to unseen environments.}
    \item {The novel \textbf{Goal Dreamer} module equips the agent with the ability to make action decisions based on the imagination of the unseen destination. 
    This module 
    further boosts the action prediction accuracy.}
    \item {
    Analyzing different codebook room types shows that visual descriptors of the room concept better generalize than textual descriptions and classification heads. This indicates that, at least in this embodied AI task, 
    visual features are 
    more informative.}
\end{itemize}

\section{Related work}
\subsubsection{Embodied Referring Expression Grounding.} In the Embodied Referring Expression Grounding task~\cite{reverie,zhu2021soon}, 
%
many prior works focus on adapting multimodal pre-trained networks to the reinforcement learning pipeline of navigation~\cite{reverie,hop} or introducing pretraining strategies for good generalization ~\cite{Qiao2022HOP,Hong_2021_CVPR}. Some recent breakthroughs come from including on-the-fly construction of a topological map and a trajectory memory as done in VLN-DUET~\cite{vlnduet}. Previous models only consider the observed history when predicting the next step. Different from them, we design a novel model to imagine future destinations while constructing the topological map. 

\subsubsection{Map-based Navigation.} In general language-guided navigation tasks, online map construction gains increasing attention (e.g., ~\cite{chaplot2020object,chaplot2020learning,irshad2022sasra}). A metric map contains full semantic details in the observed space and precise information about navigable areas. 
Recent works 
focus on improving subgoal identification~\cite{min2021film,blukis2022persistent,song2022one} and path-language alignment~\cite{wang2022find}. However, online metric map construction is inefficient during large-scale training, and its quality suffers from sensor noise in real-world applications. Other studies focus on topological maps~\cite{nrns,vlnce_topo,vlnduet}, which provide a sparser map representation and good backtracking properties. 
We use topological maps as the agent's memory. 
Our agent learns layout-aware topological node embeddings that are driven by the prediction of room type as the auxiliary task, pushing it to include commonsense knowledge of typical layouts in the representation.

\subsubsection{Visual Common Sense Knowledge.} Generally speaking, visual common sense refers to knowledge that frequently appears in a day-to-day visual world. It can take the form of a hand-crafted knowledge graph such as Wikidata~\cite{vrandevcic2014wikidata} and ConceptNet~\cite{liu2004conceptnet}, or it can be extracted from a language model~\cite{acl2022-holm}. However, the knowledge captured in these resources is usually abstract and hard to align with objects mentioned in free 
language. 
Moreover, if, for instance, you would like to know what a living room looks like, then several images of different living rooms will form a more vivid description than its word definition. 
Existing work~\cite{xiaomi@concept} 
tries to boost the agent's performance using an external knowledge graph in the Grounding Remote Referring Expressions task. 
%
Inspired by the recent use of prompts 
~\cite{petroni2019lama,brown2020gpt3} to extract knowledge from large-scale pre-trained language models (PLM)~\cite{devlin2019bert,brown2020gpt3,kojima2022large}, we consider pre-trained text-to-image models~\cite{glide,ramesh2022dalle2} as our visual commonsense resources. 
Fine-tuning a pre-trained vision-language model has been used in multimodal tasks~\cite{lu2019vilbert,su2019vl}. 
However, 
considering the explicit usage of prompted images as visual common sense for downstream tasks is novel. Pathdreamer~\cite{koh2021pathdreamer} proposes a model that predicts future observations given a sequence of previous panoramas along the path. It is applied in a VLN setting requiring detailed instructions for path sequence scoring. Our work studies the role of general visual commonsense knowledge and focuses on room-level imagination and destination imagination when dealing with high-level instructions. The experiments show that, on the one hand, including visual commonsense knowledge essentially improves task performance. On the other hand, visual common sense performs better than text labels on both environmental layout prediction and destination estimation.

\section{Methodology}

\subsection{Overview}
\begin{figure*}[t]
  \begin{center}
    \includegraphics[width=1.0\textwidth]{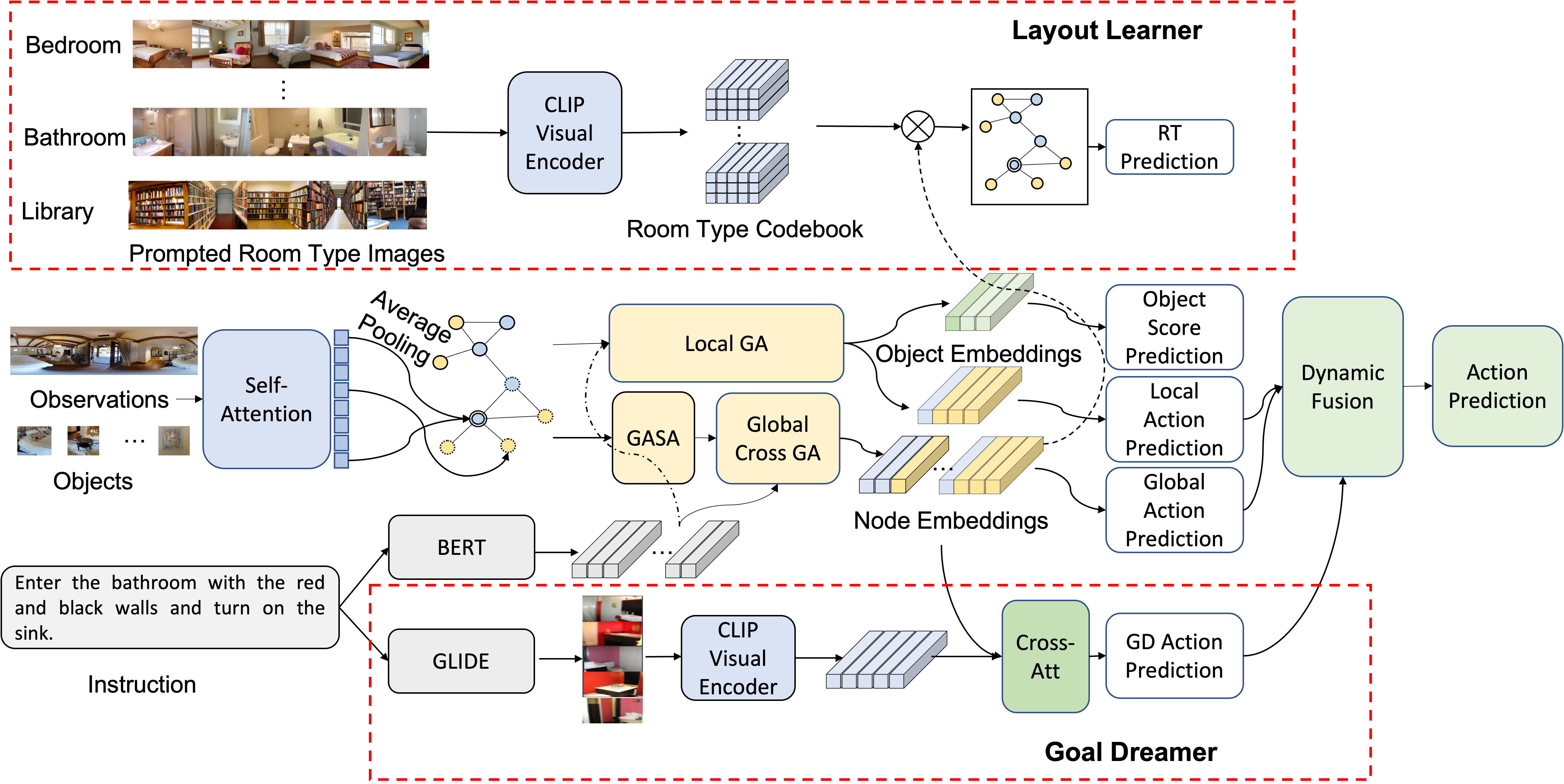}
  \end{center}
  \caption{The model architecture of our Layout-aware Dreamer (LAD). Our model predicts the room type of all nodes of the topological graph; for simplicity, we only show the predictions of several nodes here. The center part is the baseline model, which takes the topological graph and instruction as inputs and dynamically fuses the local and global branch action decisions to predict the next action. The dashed boxes show our proposed Layout Learner and Goal Dreamer.}
  \label{fig: layout}
\end{figure*}

\textbf{Task Setup.}
In Embodied Referring Expression Grounding task~\cite{reverie,zhu2021soon}, an agent is spawned at an initial position in an indoor environment. The environment is represented as an undirected graph $\mathcal{G} = (\mathcal{V}, \mathcal{E})$, where $\mathcal{V}$ stands for navigable nodes and $\mathcal{E}$ denotes connectivity edges. The agent is required to correctly localize a remote target object described by a high-level language instruction. Specifically, at the start position of each episode, the agent receives a concise high-level natural language instruction $\mathcal{X}=<w_1, w_2,..., w_L>$, where $L$ is the length of the instruction, and $w_i$ represents the $i$th word token. 
The panoramic view $\mathcal{V}_t=\{v_{t,i}\}_{i=1}^{36}$ of the agent location at time step $t$ is represented by $36$ images which are the observations of the agent with 
12 heading angles and 3 elevations. At each time step $t$, 
the agent has access to the state information $S_t$ of its current location consisting of panoramic view $\mathcal{V}_t$,
and $M$ neighboring navigable nodes $A_{t} = [a_{t,1}, ..., a_{t,M}]$ with only a single view for each of these, namely the view observable from the current node. These single views of neighboring nodes form $\mathcal{\mathcal{N}}_t=[v_{t,1}, ..., v_{t,M}]$.
%
Then the agent is required to take a sequence of actions $<a_0, ... a_t, ...>$ to reach the goal location and ground the object specified by the instruction in the observations. The possible actions $a_{t}$ at each time step $t$ are either selecting one of the navigable nodes $A_t$ or stopping at the current location denoted by $a_{t,0}$.

\subsection{Base architecture}
Our architecture is built on the basis of the VLN-DUET~\cite{vlnduet} model, which is the previous state-of-the-art model on the REVERIE dataset. In the following paragraphs, we briefly describe several important components of this base architecture, including topological graph construction, the global and local cross-attention modules with minimal changes. For more details, we refer the reader to~\citet{vlnduet}.

\subsubsection{Topological Graph.} The base model gradually builds a topological graph $\mathcal{G}_t = \{\boldsymbol v,\boldsymbol e \mid \boldsymbol v \subseteq \mathcal{V}, \boldsymbol e \subseteq \mathcal{E} \}$ to represent the environment at time step $t$. The graph contains three types of nodes: (1) visited nodes; (2) navigable nodes; and (3) the current node. In Fig.~\ref{fig: layout}, they are denoted by a blue circle, yellow circle, and double blue circle, respectively. Both visited nodes and the current node have been explored and the agent has access to their panoramic views. The navigable nodes are unexplored and can be partially observed from other visited nodes. When navigating to a node $a_{t,0}$ at time step $t$, the agent extracts panoramic image features $\mathcal{R}_t$ from its panoramic view $\mathcal{V}_t$ and object features $\mathcal{O}_t$ from provided object bounding box. The model then uses a multi-layer transformer with self-attention to model the relation between the image features $\mathcal{R}_t$ and the object features $\mathcal{O}_t$. The fused features $\mathcal{R}_t$ and $\mathcal{O}_t$ are treated as local visual features of node $a_{t,0}$. During exploring the environment, the agent updates the node visual representations as follows: (1) For the current node, the node representation is updated by concatenating and average pooling the local features $\hat{\mathcal{R}}_t$ and $\hat{\mathcal{O}_t}$. (2) As unvisited nodes could be partially observed from different directions of multiple visited nodes, the average of all image features of the partial views are taken as its representation. (3) The features of visited nodes remain unchanged. The final representation of nodes is the sum of location embedding, step embedding and visual embedding.  The location embedding of a node is formed by the concatenation of Euclidean distance, heading and elevation angles relative to the current node. The step embedding embeds the last visited time step of each visited node, and time zero is set for unvisited nodes. 

\noindent\textbf{Language Encoder.} We use a multi-layer transformer encoder~\cite{transformer} to encode the natural language instruction $\mathcal{X}$. Following the convention, we feed the sum of the token embedding, position embedding and token type embedding into the transformer encoder, and the output denoted as $\mathcal{T}$ is taken as language features.

\noindent\textbf{Global Node Self-Attention.} Different from the VLN-DUET model, to enable each node to perceive global environment information without influenced by the language information, we conduct one more graph aware self-attention (GASA) \cite{vlnduet} over node embeddings $\mathcal{H}_t$ of graph $\mathcal{G}_t$ before interacting with word embeddings. For simplicity, we use the same symbol $\mathcal{H}_t$ to denote the encoded graph node embeddings.


\noindent\textbf{Cross Graph Encoder.} Following the work of~\citet{devlin2019bert}, we use a multimodal transformer~\cite{lu2019vilbert} to model both the global and local graph-language relation. We name the global and local graph-language cross-attention models (Global Cross GA and Local GA) as global branch and local branch, respectively. For the global branch, we perform cross-attention of node embeddings $\mathcal{H}_t$ over language features $\mathcal{T}$, while only the current node and its neighboring navigable nodes are used to compute the cross-attention in the local branch. We feed the outputs of the global branch to the \text{Layout Learner} for layout prediction. In addition, both the global and local branch outputs are further used to make the navigation decision, as shown in Fig.~\ref{fig: layout}. 

\begin{align*}
    \mathcal{\tilde{H}}_t^{(glo)} &= \text{Cross-Attn}(\mathcal{H}_t, \mathcal{T}, \mathcal{T})  \tag{1} \\
    \mathcal{\tilde{H}}_t^{(loc)} &= \text{Cross-Attn}(\{\mathcal{H}_t(\mathcal{A}_t), \mathcal{H}_t(a_{t,0})\}, \mathcal{T}, \mathcal{T}) \tag{2}
\end{align*}
where $\text{Cross-Attn}(query, key, value)$
is a multi-layer transformer decoder, and $\mathcal{A}_t$ stands for neighbouring navigable nodes of the current node $a_{t,0}$. $\mathcal{H}_t(\cdot)$ represents extracting corresponding rows from $\mathcal{H}_t$ by node indices. 

\subsection{Layout Learner}
This module aims to learn both the implicit environment layout distribution and visual commonsense knowledge of the room type, which is achieved through an auxiliary layout prediction task with our room type codebook. This auxiliary task is not used directly at inference time. The main purpose of having it during training is learning representations to capture this information, which in turn improves global action prediction.

\begin{figure}[t]
    \centering
    \includegraphics[width=\linewidth]{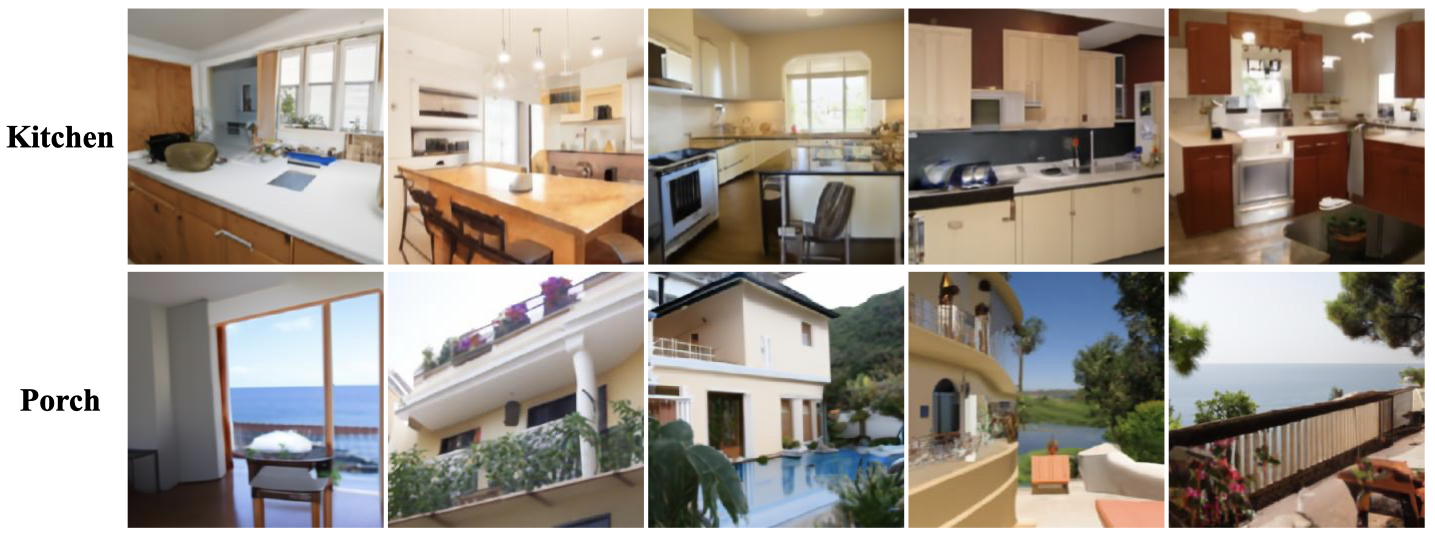}
    \caption{Prompted examples of the room codebook. }
    \label{fig: glide room}
\end{figure}

\noindent\textbf{Building Room Type Codebook.}
We fetch room type labels from the MatterPort point-wise semantic annotations which contain 30 distinct room types.
We then select the large-scale pre-trained text-to-image generation model GLIDE~\cite{glide} as a visual commonsense resource. To better fit the embodied grounding task, we create prompt $P_{room}$ to prompt visual commonsense knowledge not only based on the room type label but including high-frequency objects of referring expressions in the training set. Specifically, when building the room type codebook, we create prompts by filling in the following template.
\begin{align*}
    \text{A [room type] with [obj 1], ... and [obj n]}.
\end{align*}
where [room type] is a room label annotated in Matterport3D dataset~\cite{Matterport3D} with manual spelling completion, such as map ``l" to ``living room" and ``u" to ``utility room". [obj 1] .. [obj n] are high-frequency object words that co-occur with specific room labels in the training instructions of the REVERIE dataset. A frequency above 0.8 is considered a high frequency. This threshold ensures diversity and limits the number of candidates. For instance, we create the prompt ``A dining room with table and chairs" for room type ``dining room", ``a bathroom with towel and mirror" for room type ``bathroom". For each room type, we generate a hundred images and select $S$ representative ones in the pre-trained CLIP feature space (i.e., the image closest to each clustering center after applying a K-Means cluster algorithm). An example is shown in Fig.~\ref{fig: glide room}. Our selection strategy guarantees the diversity of the generated images, i.e., rooms from various perspectives and lighting conditions. The visual features of the selected images for different room types form the room type codebook $E_{room}\in \mathbb{R}^{K \times S \times 765}$, where $K$ is the total number of room types and $S$ represents the number of images for each room type. This codebook represents a commonsense knowledge base with visual descriptions of what a room should look like.

\noindent\textbf{Environment Layout Prediction.} Layout information is critical for indoor navigation, especially when the agent only receives high-level instructions describing the goal positions, such as ``Go to the kitchen and pick up the mug beside the sink". This module equips the agent with both the capability of learning room-to-room correlations in the environment layout and a generalized room type representation. With the help of the visual commonsense knowledge of the rooms in the room type codebook, we perform layout prediction. 
We compute the similarity score between node representations $\mathcal{\tilde{H}}^{(glo)}_t$ and image features $E_{room}$ in the room type codebook and further use this score to predict the room type of each node in the graph $\mathcal{G}_t$. The predicted room type logits are supervised with ground truth labels $\mathcal{C}_t$.
\begin{align*}
    \hat{\mathcal{C}_{t}^{i}} &= \sum_{j=1}^{S}{\mathcal{\tilde{H}}^{(glo)}_t E_{room (i,j)}} \tag{3} \\
    \mathcal{L}_t^{\text{(LP)}} &=\text{CrossEntropy}(\hat{\mathcal{C}_t},\mathcal{C}_t) \tag{4}
\end{align*}
where $S$ is the number of images in the room type codebook for each room type, and $\hat{\mathcal{C}}_t^i$ represents the predicted score of $i$th room type. We use $\hat{\mathcal{C}_t}$ to denote the predicted score distribution of a node, thus $\hat{\mathcal{C}_t} = [\hat{\mathcal{C}_t^0},\cdots,\hat{\mathcal{C}_t^{K}}]$.

\subsection{Goal Dreamer} 
\begin{figure}[H]
    \centering
    \includegraphics[width=0.95\linewidth]{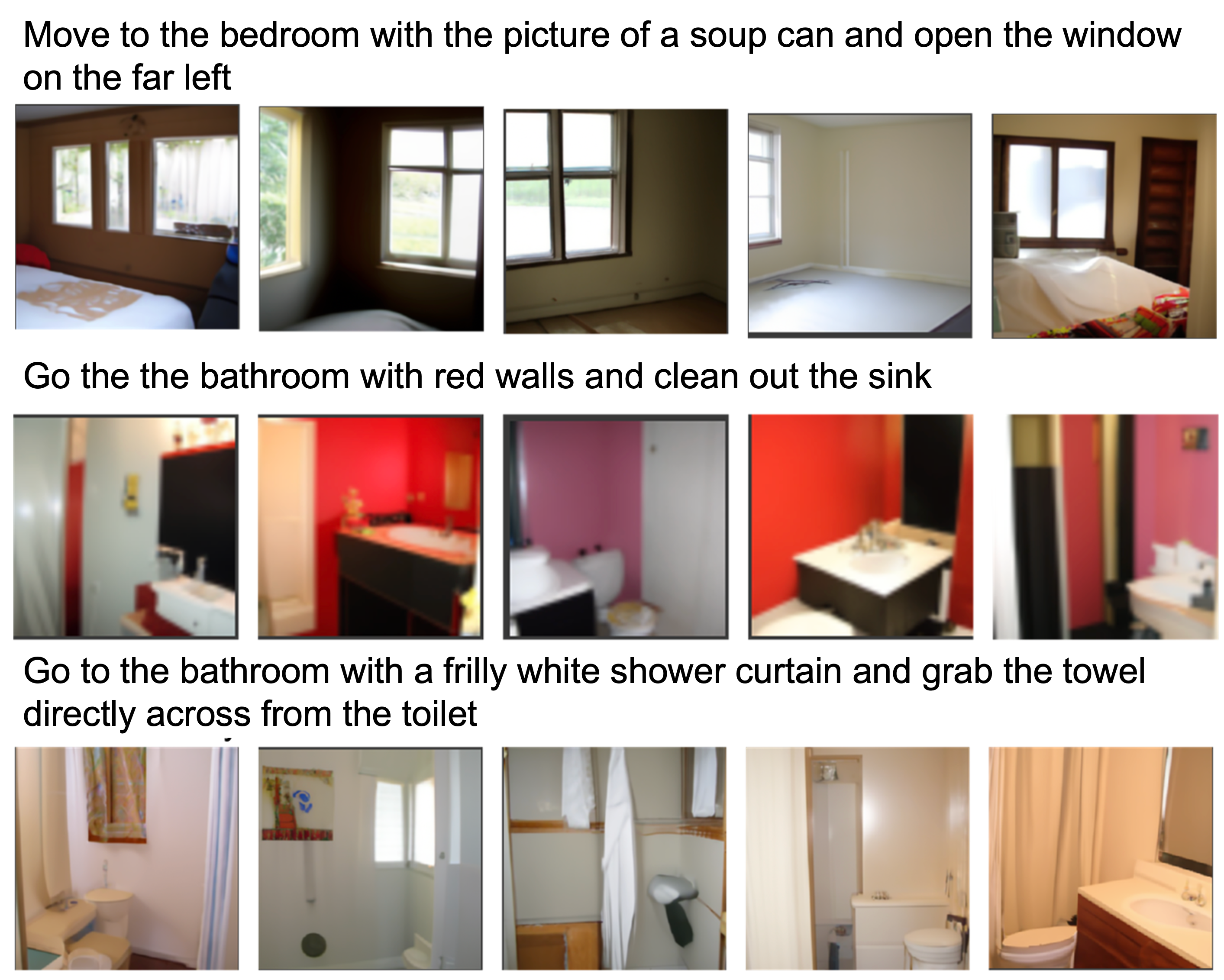}
    \caption{Images of the destination 
    generated by GLIDE 
    based on the given instruction.}
    \label{fig: glide des}
\end{figure}

A navigation agent without a global map can be short-sighted. We design a long-horizon value function to guide the agent toward the imagined destination. For each instruction, we prompt five images from GLIDE~\cite{glide} as the imagination of the destination. Three examples are shown in Fig~\ref{fig: glide des}. Imagination features $E^{(im)}$ are extracted from the pre-trained CLIP vision encoder~\cite{clip}. Then at each time step $t$, we attend the topological global node embeddings $\mathcal{\tilde{H}}_t^{(glo)}$ to $E^{(im)}$ through a cross-attention layer~\cite{transformer}.

\begin{align*}
    \mathcal{\hat{H}}_t^{(glo)} = \text{Cross-Attn}(\mathcal{\tilde{H}}_t^{(glo)}, E^{(im)}, E^{(im)}) \tag{5}
\end{align*}
The hidden state $\mathcal{\hat{H}}_t^{(glo)}$ learned by the Goal Dreamer is projected by a linear feed-forward network (FFN)\footnote{FFNs in this paper are independent without parameter sharing.} to predict the probability distribution of the next action step over all navigable but not visited nodes.
\begin{align*}
 Pr_{t}^{(im)} = \text{Softmax}(\text{FFN}(\mathcal{\hat{H}}_t^{(glo)}))\tag{6}
\end{align*}

We supervise this distribution $Pr_{t}^{(im)}$ in the warmup stage of the training (see next Section) with the ground truth next action $\mathcal{A}_{gt}$.
\begin{align*}
 \mathcal{L}_t^{(D)} &= \text{CrossEntropy}(Pr_{t}^{(im)},\mathcal{A}_{gt}) \tag{7}
\end{align*}
Optimizing $Pr_{t}^{(im)}$ guides the learning of latent features
$\mathcal{\hat{H}}_t^{(glo)}$. $\mathcal{\hat{H}}_t^{(glo)}$ will be fused with global logits in the final decision process as described in the following section.


\subsection{Decision Maker}
\noindent\textbf{Action Prediction}.
We follow the work of~\citet{vlnduet} to predict the next action to be performed in both the global and local branches and dynamically fuse their results to enable the agent to backtrack to previous unvisited nodes.
\begin{align*}
    Pr_{t}^{\text{(floc)}} = \text{DynamicFuse}(\mathcal{\tilde{H}}_t^{(loc)}, \mathcal{\tilde{H}}_t^{(glo)}) \tag{8}
\end{align*}
The proposed goal dreamer module equips the agent with the capability of learning guidance 
towards the target goal, hence we further fuse the goal dreamer's latent features $\mathcal{\hat{H}}_t^{(glo)}$ with global results weighted by a learnable $\lambda_t$. The $\lambda_t$ is node-specific; thus we apply a feed-forward network (FFN) to predict these weights conditioned on node representations. 
\begin{align*}
    \lambda_t &= \text{FFN}([\mathcal{\tilde{H}}_t^{(glo)};\mathcal{\hat{H}}_t^{(glo)}]) \tag{9}
\end{align*}
The fused action distribution is formulated as:
\begin{align*}
    Pr_{t}^{\text{(fgd)}} &= (1-\lambda_t) * \mathcal{\text{FFN}(\tilde{H}}_t^{(glo)}) \\
    &+ \lambda_t * \mathcal{\text{FFN}(\hat{H}}_t^{(glo)})  \tag{10}
\end{align*}
The objective for supervising the whole decision procedure by ground truth node $\mathcal{A}_{gt}$ in the next time step is: 
\begin{align*}
Pr_t^{\text{(DSAP)}} &= Pr_{t}^{\text{(floc)}} + Pr_{t}^{\text{(fgd)}} \\
\mathcal{L}_t^{\text{(DSAP)}} &= \text{CrossEntropy}(Pr_t^{\text{(DSAP)}} ,\mathcal{A}_{gt}) \tag{11}
\end{align*}
where $Pr_t^{\text{(DSAP)}}$ is the estimated single-step prediction distribution (DSAP) over all nodes. Not only the global-local fusion already proposed in previous work but also our goal-dreaming branch is now included to predict the next action.

\noindent\textbf{Object Grounding}.
We simply consider object grounding as a classification task and use a $\text{FFN}$ to generate a score for each object in $\mathcal{O}_t$ of the current node. We then supervise this score with the annotated ground truth object $\mathcal{O}_{gt}$. 
\begin{align*}
    \hat{\mathcal{O}_t} &= \text{FFN}(\mathcal{O}_t) \\ 
    \mathcal{L}_t^{\text{(OG)}} &= \text{CrossEntropy}(\hat{\mathcal{O}_t}, \mathcal{O}_{gt}) \tag{12}
\end{align*}

\subsection{Training and Inference}
\label{sec: training}
\textbf{Warmup stage.} Previous researches~\cite{history,vlnduet,episodic,hao2020towards} have shown that warming up the model with auxiliary supervised or self-supervised learning tasks can significantly boost the performance of a transformer-based VLN agent. We warm up our model with five auxiliary tasks, including three common tasks in vision-and-language navigation: masked language modeling (MLM)~\cite{devlin2019bert}, masked region classification (MRC)~\cite{lu2019vilbert}, object grounding (OG)~\cite{hop} if object annotations exist; and two new tasks, that is, layout prediction (LP) and single action prediction with the dreamer (DSAP) explained in sections \text{Layout Learner} and \text{Goal Dreamer}, respectively. In the \text{LP}, our agent predicts the room type of each node in the topological graph at each time step, aiming to model the room-to-room transition of the environment, the objective $\mathcal{L}_t^{\text{(LP)}}$ of which is shown in Eq. 4. To encourage the agent to conduct goal-oriented exploration, in the \text{DSAP}, as illustrated in $\mathcal{L}_t^{\text{(D)}}$ (Eq. 7) and $\mathcal{L}_t^{\text{(DSAP)}} $ (Eq. 11) we use the output of the goal dreamer to predict the next action.
The training objective of the warmup stage is as follows:
\begin{align*}
    \mathcal{L}^{\text{WP}}_t = &\mathcal{L}_t^{\text{(MLM)}} +\mathcal{L}_t^{\text{(MRC)}} + \mathcal{L}_t^{\text{(OG)}} + \\
    & \mathcal{L}_t^{\text{(LP)}} + \mathcal{L}_t^{\text{(D)}} + \mathcal{L}_t^{\text{(DSAP)}}  \tag{13}
\end{align*}

\noindent\textbf{Imitation Learning and Inference.} We use the imitation learning method DAgger~\cite{dagger} to further train the agent. During training, we use the connectivity graph $\mathcal{G}$ of the environment to select the navigable node with the shortest distance from the current node to the destination as the next target node. We then use this target node to supervise the trajectory sampled using the current policy at each iteration. 
The training objective here 
is:
\begin{align*}
    \mathcal{L}^{\text{IL}}_t = & \mathcal{L}_t^{\text{(OG)}} + \mathcal{L}_t^{\text{(LP)}} + \mathcal{L}_t^{\text{(DSAP)}}   \tag{14}
\end{align*}
During inference, our agent builds the topological map on-the-fly and selects the action with the largest probability. If the agent decides to backtrack to the previous unexplored nodes, the classical Dijkstra algorithm~\cite{dijkstra1959note} is used to plan the shortest path from the current node to the target node. The agent stops either when a stop action is predicted at the current location or when it exceeds the maximum action steps. When the agent stops, it selects the object with the maximum object prediction score.

\section{Experiments}
\subsection{Datasets}
Because 
the navigation task is characterized by realistic 
high-level instructions, we conduct experiments and evaluate our agent on the embodied goal-oriented benchmark REVERIE~\cite{reverie} and 
the SOON~\cite{song2022one} datasets.

REVERIE dataset: The dataset is split into four sets, including the training set, validation seen set, validation unseen set, and test set. The environments in both validation unseen and test sets do not appear in the training set, while all environments in validation seen have been explored or partially observed during training. The average length of instructions is $18$ words. The dataset also provides object bounding boxes for each panorama, and the length of ground truth paths ranges from $4$ to $7$ steps.

SOON dataset: This dataset has a similar data split as REVERIE. The only difference is that it proposes a new validation on the seen instruction split which contains the same instructions in the same house but with different starting positions. Instructions in SOON contain $47$ words on average, and the ground truth paths range from $2$ to $21$ steps with $9.5$ steps on average. The SOON dataset does not provide bounding boxes for object grounding, thus we use here an existing object detector~\cite{anderson2018bottom} to generate candidate bounding boxes.

\subsection{Evaluation Metrics}
\noindent\textbf{Navigation Metrics.} Following previous work~\cite{vlnduet,evaluation}, we evaluate the navigation performance of our agent using standard 
metrics, including Trajectory Length (TL) which is the average path length in meters; Success Rate (SR) defined as the ratio of paths where the agent's location is less than $3$ meters away from the target location; Oracle SR (OSR) that defines success if the trajectory has ever passed by the target location; and SR weighted by inverse Path Length (SPL).

\noindent\textbf{Object Grounding Metrics.} We follow the work of~\citet{reverie} using 
Remote Grounding Success (RGS), which is the ratio of successfully executed instructions, and RGS weighted by inverse Path Length (RGSPL). 

\subsection{Implementation Details.} 
The model is trained for 100k iterations with a batch size of 32 for single action prediction and 50k iterations with a batch size of 8 for imitation learning with DAgger~\cite{dagger}. We optimize both phases by the AdamW~\cite{adamw} optimizer with a learning rate of 5e-5 and 1e-5, respectively. We include two fixed models for preprocessing data, i.e., GLIDE~\cite{glide} for generating the room codebook and the imagined destination, and CLIP~\cite{clip} for image feature extraction. 
The whole training procedure takes two days with a single NVIDIA-P100 GPU.

\section{Results}

\begin{table*}[t]
    \centering
    \resizebox{\linewidth}{!}{
    \begin{tabular}{l|cccc|cc||cccc|cc||cccc|cc}
    \hline 
    \multirow{3}*{Methods} & \multicolumn{6}{c}{Val-seen} & \multicolumn{6}{c}{Val-unseen} & \multicolumn{6}{c}{Test-unseen} \\
    ~ & \multicolumn{4}{c}{Navigation} & \multicolumn{2}{c}{Grounding} &\multicolumn{4}{c}{Navigation} &
    \multicolumn{2}{c}{Grounding} &\multicolumn{4}{c}{Navigation} &
    \multicolumn{2}{c}{Grounding} \\ 
    \hline
     ~ & TL $\downarrow$ & OSR$\uparrow$ & SR$\uparrow$ & SPL$\uparrow$ & RGS$\uparrow$ & RGSPL$\uparrow$ & TL $\downarrow$ & OSR$\uparrow$ & SR$\uparrow$ & SPL$\uparrow$ & RGS$\uparrow$ & RGSPL$\uparrow$ & TL $\downarrow$ & OSR$\uparrow$ & SR$\uparrow$ & SPL$\uparrow$ & RGS$\uparrow$ & RGSPL$\uparrow$ \\
     \hline 
     RCM~\cite{reinforced} & 10.70 & 29.44 & 23.33 & 21.82 & 16.23 & 15.36 & 11.98 & 14.23 & 9.29 & 6.97 & 4.89 & 3.89 & 10.60 & 11.68 & 7.84 & 6.67 & 3.67 & 3.14 \\
     SelfMonitor~\cite{ma2019self} & 7.54 & 42.29 & 41.25 & 39.61 & 30.07 & 28.98 & 9.07 & 11.28 & 8.15 & 6.44 & 4.54 & 3.61 & 9.23 & 8.39 & 5.80 & 4.53 & 3.10 & 2.39 \\ 
     REVERIE~\cite{reverie} & 16.35 & 55.17 & 50.53 & 45.50 & 31.97 & 29.66 & 45.28 & 28.20 & 14.40 & 7.19 & 7.84 & 4.67 & 39.05 & 30.63 & 19.88 & 11.61 & 11.28 & 6.08 \\
     CKR~\cite{gao2021room} & 12.16 & 61.91 & 57.27 & 53.57 & 39.07 & - & 26.26 & 31.44 & 19.14 & 11.84 & 11.45& - & 22.46 & 30.40 & 22.00 & 14.25 & 11.60 & - \\
     SIA~\cite{hop} & 13.61 & 65.85 & 61.91 & 57.08 & 45.96 & 42.65 & 41.53 & 44.67 &  31.53 & 16.28 & 22.41 & 11.56 & 48.61 & 44.56 & 30.8 & 14.85 & 19.02 & 9.20 \\
     VLN-DUET~\cite{vlnduet} & 13.86 & \textbf{73.86} & \textbf{71.75} & \textbf{63.94} &  \textbf{57.41} & \textbf{51.14} & 22.11 & 51.07 & 46.98 & 33.73 & 32.15 & 22.60 & 21.30 & 56.91 & 52.51 & 36.06 & 31.88 & 22.06 \\ 
    \hline
    \textbf{LAD (Ours)} &  16.74 & 71.68 & 69.22 & 57.44 & 52.92 & 43.46 & 26.39 & \textbf{63.96} & \textbf{57.00} & \textbf{37.92} & \textbf{37.80} & \textbf{24.59} & 25.87 & \textbf{62.02} & \textbf{56.53} & \textbf{37.8} & \textbf{35.31} &  \textbf{23.38} \\ 
    \hline  
    \end{tabular}
    }
    \caption{Results obtained on the REVERIE dataset as compared to other existing models including the current state-of-the-art model VLN-DUET. 
    }
    \label{tab: main table}
\end{table*}

\subsection{Comparisons to the state of the art.}
\noindent\textbf{Results on REVERIE.} 
In Table~\ref{tab: main table}, we compare our model with prior works in four categories: (1) Imitation Learning + Reinforcement learning models: RCM~\cite{reinforced}, SIA~\cite{hop}; (2) Supervision model: SelfMonitor~\cite{ma2019self}, REVERIE~\cite{reverie}; (3) Imitation Learning with external knowledge graph: CKR~\cite{gao2021room}; and (4) Imitation Learning with topological memory: VLN-DUET~\cite{vlnduet}.
Our model outperforms the above models with a large margin in challenging unseen environments. Significantly, our model surpasses the previous state of the art 
\text{VLN-DUET} by approximately $10\%$ (\text{SR}) and $5\%$ (\text{RGS}) in the val-unseen split. On the test split, our model beats VLN-DUET with improvements of \text{SR} by $4.02\%$ and \text{RGS} by $3.43\%$. The results demonstrate that the proposed \text{LAD} better generalizes to unseen environments, which is critical for real applications.

\noindent\textbf{Results on SOON.} Table~\ref{tab: soon} presents the comparison of our proposed LAD with other models including the state-of-the-art VLN-DUET model. The LAD model significantly outperforms VLN-DUET across all evaluation metrics in the challenging test unseen split. Especially, the model improves the performance on \text{SR} and \text{SPL} by $6.15\%$ and $6.4\%$, respectively. This result clearly shows the effectiveness of the proposed \text{Layout Learner} and \text{Goal Dreamer} modules.

\subsection{Ablation Studies}
We verify the effectiveness of our key contributions via an ablation study on the REVERIE dataset.


\begin{figure*}[t]
    \centering
    \includegraphics[width=\textwidth]{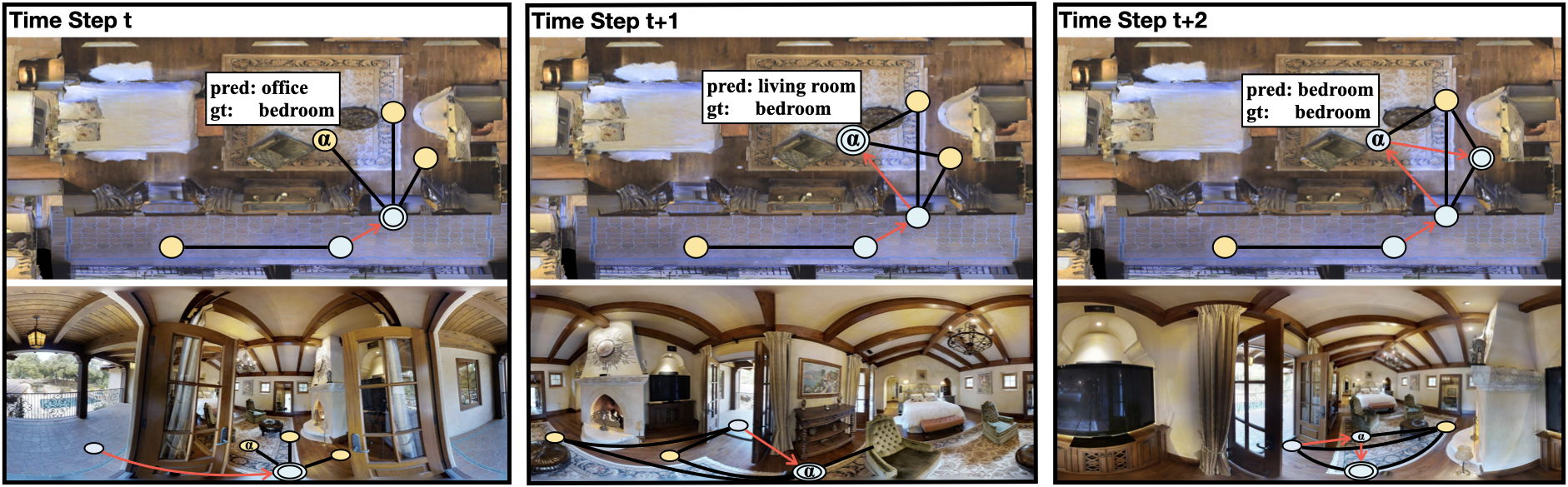}
    \caption{Room type belief will be updated while exploring. Blue double circle denotes the current location, a yellow circle refers to an unexplored but visible node, a blue circle represents a visited node, the red line is the agent's trajectory. Each column contains bird-view and egocentric views of current agent states.}
    \label{fig: room change}
    
\end{figure*}

\begin{table}[H]
    \centering
    \resizebox{\linewidth}{!}{
    \begin{tabular}{c|c|ccccc}
    \hline
    Split & Methods & TL $\downarrow$ & OSR $\uparrow$ & SR $\uparrow$ & SPL $\uparrow$ & RGSPL $\uparrow$ \\  
    \hline
    \multirow{3}*{\makecell[c]{Val \\ Unseen}} & GBE~\cite{zhu2021soon} & 28.96 & 28.54 & 19.52 & 13.34 & 1.16  \\
    ~ & VLN-DUET~\cite{vlnduet} & 36.20 & \textbf{50.91} & 36.28 & 22.58 & 3.75  \\ 
    ~ & LAD & 32.32 & 50.59 & \textbf{40.24} & \textbf{29.44} & \textbf{4.20} \\
    \hline 
    \multirow{3}*{\makecell[c]{Test \\ Unseen}} & GBE~\cite{zhu2021soon} & 27.88 & 21.45 & 12.90 & 9.23 & 0.45 \\
    ~ & VLN-DUET~\cite{vlnduet} & 41.83 & 43.00 & 33.44 & 21.42 & 4.17 \\
    ~ & LAD  & 35.71 & \textbf{45.80} & \textbf{39.59} &\textbf{27.82} & \textbf{7.08}   \\ 
    \hline 
    \end{tabular}
    }
    \caption{Results of our LAD model obtained on the SOON dataset compared with the results of other state-of-the-art models.
    }
    \label{tab: soon}
\end{table}

\noindent\textbf{Are the Layout Learner and Goal Dreamer helpful? } We first verify the contribution of the \text{Layout Learner} and \text{Goal Dreamer}. For a fair comparison, we re-implement the result of VLN-DUET~\cite{vlnduet} but replace the visual representations to CLIP features and report the results in Row 1 of Table~\ref{tab: abla_table}. The performance boost in this re-implementation compared to VLN-DUET results in Table~\ref{tab: main table} indicates that CLIP features are more suitable for visual-language tasks than original ImageNet ViT features. Comparing the results in row $2$ with row $1$, it is clear that integrating the \text{Layout Learner} to the baseline model improves its performance across all evaluation metrics, which verifies our assumption that the layout information is vital in high-level instruction following tasks. One might notice that in row $3$ the \text{Goal Dreamer} module can boost the performance in \text{SR}, \text{OSR}, \text{SPL}, and \text{RGSPL}, but it slightly harms the performance in \text{RGS}. 
A lower \text{RGS} but higher \text{RGSPL} shows that the model with \text{Goal Dreamer} takes fewer steps to reach the goal, meaning that it conducts more effective goal-oriented exploration, which supports our assumption.

\begin{table}[t]
\resizebox{\linewidth}{!}{
\begin{tabular}{ccc|ccccc}
\hline 
Baseline & Layout Learner & Goal Dreamer & OSR$\uparrow$ & SR$\uparrow$ & SPL$\uparrow$ & RGS$\uparrow$ & RGSPL$\uparrow$ \\ 
\hline 
\checkmark&  & & 58.68 & 52.34 & 34.45 & 35.02 & 22.87 \\
\checkmark & \checkmark &  & 63.90 & 56.04 & 37.66 & 37.06 & 24.58 \\
\checkmark &  & \checkmark & 61.03 & 53.45 & 37.41 & 34.34 & 24.03 \\ 
\checkmark & \checkmark & \checkmark & 63.96 & 57.00 & 37.92 & 37.80 & 24.59 \\ 
\hline 
\end{tabular}
}
\caption{Comparisons of baseline model and baseline with our proposed modules (Layout Learner and  Goal Dreamer).}
\label{tab: abla_table}
\end{table}

\noindent\textbf{Visual or textual common sense?} 
In this work, we consider several images to describe a commonsense concept. 
In this experiment, we study whether visual descriptors of room types lead to a better generalization than directly using the classification label or a textual description while learning an agent. 
In the first line of Table~\ref{tb: codeboook}, we show the results of directly replacing the visual codebook module with a room label classification head. It shows a $3\%$ drop in both navigation and grounding success rates. This indicates that a single room type classification head is insufficient for learning good latent features of room concepts.
We further compare the results of using a visual codebook with using a textual codebook. Since we use text to prompt multiple room images as our visual room codebook 
encoded with the CLIP~\cite{clip} visual encoder, for a fair comparison, we encode the text prompts as a textual codebook using the CLIP text encoder. Then we replace the visual codebook in our model with the textual one and re-train the whole model. As shown in Table \ref{tb: codeboook}, the textual codebook has a $5.62\%$ drop in navigation success rate (SR) and a $5.58\%$ drop in remote grounding success rate (RGS). This indicates that visual descriptors of commonsense room concepts are informative and easier to follow for an autonomous agent.
\begin{table}[t]
\resizebox{\linewidth}{!}{
\begin{tabular}{ccc|ccccc}
\hline 
FFN & Text & Visual & OSR$\uparrow$ & SR$\uparrow$ & SPL$\uparrow$ & RGS$\uparrow$ & RGSPL$\uparrow$ \\
\hline
\checkmark & &  &58.76 &53.73 & 35.22 &34.68 & 23.58 \\
&\checkmark & &  56.06 & 51.38 & 35.57 & 32.38 & 22.05 \\ 
& & \checkmark  & 63.96 & 57.00 & 37.92 & 37.80 & 24.59 \\
\hline 
\end{tabular}
}
\caption{Codebook type comparison: visual room codebook versus textual room codebook and direct classification head.}
\label{tb: codeboook}
\end{table}

\noindent\textbf{Could the room type prediction be corrected while exploring more?}
In this section, we study the predicted trajectory. As shown in Fig.~\ref{fig: room change}, the incorrect room type prediction of node $\alpha$ is corrected after exploration of the room. At time step $t$, the observation only contains chairs, the prediction of room type of node $\alpha$ is office. When entering the room at time step $t+1$, the table and television indicate this room is more likely to be a living room. While grabbing another view from a different viewpoint, the room type of node $\alpha$ is correctly recognized as a bedroom. Since the instruction states to find the pillow inside the bedroom, the agent could correctly track its progress with the help of room type recognition and successfully execute the instruction. This indicates that the ability to correct former beliefs benefits the layout understanding of the environment and further has a positive influence 
on the action decision process. We further discuss the room type correction ability quantitatively. The following Fig.~\ref{fig: room acc} shows the room type recognition accuracy w.r.t. time step $t$ in the validation unseen set of the REVERIE dataset. 
It shows that room type recognition accuracy increases with increased exploration of the environment.  
We also observe that the overall accuracy of the room type recognition is still not satisfactory. We assume the following main reasons: first, room types defined in MatterPort3D have ambiguity, such as family room and living room do not have a well-defined difference; second, many rooms do not have a clear boundary in the visual input (no doors), so it is hard to distinguish connected rooms from the observations. These ambiguities require softer labels while learning, which is also a reason why using images as commonsense resource performs better than using textual descriptors and linear classification heads as is seen in Table~\ref{tb: codeboook}.

\section{Limitations and future work}
In this paper, we describe our findings while including room type prediction and destination imagination 
in the Embodied Referring Expression Grounding task, but several limitations still require further study. 

\noindent\textbf{Imagination is not dynamic} 
and it is only conditioned on the given instruction. Including observations and dynamically modifying the imagination with a trainable generation module could be helpful for fully using the knowledge gained during exploration. This knowledge could guide the imagination model to generate destination images of a style similar to the environment. It is also possible to follow the idea of PathDreamer~\cite{koh2021pathdreamer} and Dreamer~\cite{hafner2019dream,hafner2020dreamer2}, which generate a sequence of hidden future states based on the history to enhance reinforcement learning models.

\noindent\textbf{Constant number of generated visual features.} Due to the long generation time and storage consumption, we only generate five images as the goal imaginations. It is possible to increase diversity by generating more images. Then, a better sampling strategy for the visual room codebook construction and destination imagination could be designed, such as randomly picking a set of images from the generated pool. Since we have observed overfitting in the later stage of the training, it is possible to further improve the generalization of the model by including randomness in this way.

\begin{figure}
    \centering
    \includegraphics[width=\linewidth]{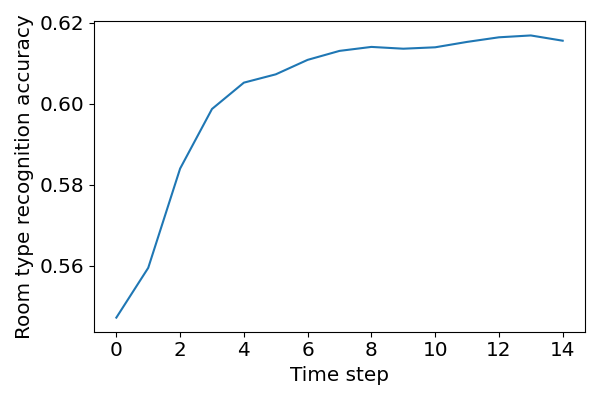}
    \caption{Room recognition accuracy of the validation unseen set of the REVERIE dataset.}
    \label{fig: room acc}
    
\end{figure}

\section{Conclusion}
In this work, to enhance the environmental understanding of an autonomous agent in the Embodied Referring Expression Grounding task, we have proposed a Layout Learner and Goal Dreamer. These two modules effectively introduce visual common sense, in our case via an image generation model, into the decision process. Extensive experiments and case studies show the effectiveness of our designed modules. We hope our work inspires further studies that include visual commonsense resources in autonomous agent design.

\section{Acknowledgements}
This research received funding from the Flanders AI Impuls Programme - FLAIR and from the European Research Council Advanced Grant 788506.
\bibliography{aaai23}


\end{document}